\documentclass[a4paper,fleqn]{cas-dc}

\usepackage[authoryear,sort&compress]{natbib}
\usepackage{tikz}

\def\tsc#1{\csdef{#1}{\textsc{\lowercase{#1}}\xspace}}
\tsc{WGM}
\tsc{QE}
\begin{document}
\let\WriteBookmarks\relax
\def\floatpagepagefraction{1}
\def\textpagefraction{.001}

\shorttitle{Stable and Robust SLIP Model Control via Energy Conservation-Based Feedback Cancellation for Quadrupedal Applications} 

\shortauthors{M. Saud Ul Hassan, D. Vasquez, H. Asif, C. Hubicki}

\title [mode = title]{Stable and Robust SLIP Model Control via Energy Conservation-Based Feedback Cancellation for Quadrupedal Applications}

\author{Muhammad Saud Ul Hassan}[orcid=0009-0008-3542-1872]

\ead{ms18ig@fsu.edu}

\cormark[1]

\author{Derek Vasquez}[orcid=0009-0002-4002-2151]

\ead{davasquez@fsu.edu}


\author{Hamza Asif}[orcid=0000-0002-8812-3472]

\ead{muhammad1.nizami@famu.edu}

% Forth author
\author{Christian Hubicki}[orcid=0000-0002-2092-3772]

\ead{hubicki@eng.famu.fsu.edu}

\affiliation{organization={Department of Mechanical Engineering},
            addressline={Florida State University}, 
            city={Tallahassee},
            state={FL},
            citysep={}, % Uncomment if no comma needed between city and postcode
            postcode={32306}, 
            country={United States}}

\cortext[1]{Corresponding author}

\begin{abstract}
In this paper, we present an energy-conservation based control architecture for stable dynamic motion in quadruped robots. We model the robot as a Spring-loaded Inverted Pendulum (SLIP), a model well-suited to represent the bouncing motion characteristic of running gaits observed in various biological quadrupeds and bio-inspired robotic systems. The model permits leg-orientation control during flight and leg-length control during stance, a design choice inspired by natural quadruped behaviors and prevalent in robotic quadruped systems. 

Our control algorithm uses the reduced-order SLIP dynamics of the quadruped to track a stable parabolic spline during stance, which is calculated using the principle of energy conservation. Through simulations based on the design specifications of an actual quadruped robot, Ghost Robotics Minitaur\textsuperscript{\texttrademark}, we demonstrate that our control algorithm generates stable bouncing gaits. Additionally, we illustrate the robustness of our controller by showcasing its ability to maintain stable bouncing even when faced with up to a 10\% error in sensor measurements.
\end{abstract}


\begin{highlights}
\item Energy-Shaping Gait Control: Present an energy-conservation-based control system for achieving stable dynamic gaits in quadruped robots.
\item Bio-Inspired Reduced-Order Modeling: Employ the Spring-loaded Inverted Pendulum (SLIP) model for quadruped dynamics, a biologically inspired computationally efficient approach, representative of bouncing dynamics of running quadrupeds in nature.
\item Real-world Simulation: Validate the proposed control algorithm in MATLAB-based simulations based on the design parameters of Ghost Robotics Minitaur\textsuperscript{\texttrademark}, a real-world quadruped robot.
\item Robustness to Noise: Demonstrate the controller's robustness by maintaining stable bouncing even with up to 10\% error in state measurement.
\end{highlights}


\maketitle

\section{Introduction}
Quadrupeds are among nature's most adept terrestrial locomotors, exhibiting an impressive combination of speed, agility, and adaptability across a variety of terrains. From cheetahs that reach speeds of up to 75 mph, making them the fastest land animals \citep{633eac20-7e46-325f-9a8c-62fb083e9ad3}, to ibex goats that are able to navigate near-vertical slopes and rocky cliffs\citep{ibex-inproceedings}, quadrupedal locomotion in nature continues to inspire research in robotics \citep{chen2019trot,7758092,8630416,doi:10.1177/0278364918779874,ma2019first}.

A critical aspect of achieving such efficient and robust locomotion in robots lies in understanding and replicating the underlying principles of animal biomechanics. Research in animal locomotion has shown that quadrupeds often employ compliant, energy-efficient gaits that make use of their elastic tendons and muscles to conserve and recycle energy during motion \citep{doi:10.1146/annurev.ph.44.030182.000525, doi:10.1152/ajpregu.1977.233.5.R243}. Specifically, during fast gaits such as trotting and galloping, animals exhibit a "bouncing" behavior, storing energy during limb compression upon touchdown and releasing it during extension for liftoff \citep{alexander1975mechanics}, enabling rapid terrestrial locomotion while minimizing energy consumption.

\input{tikz/quad}

Mechanically, the biological mechanism of energy-storage in quadrupeds may be modeled as a spring-mass system, commonly referred to as the spring-loaded inverted pendulum (SLIP) model \citep{BLICKHAN19891217, heglund1988speed}. The SLIP model captures the essence of compliant running by abstracting the animal (or robot) as a point mass bouncing on a massless springy leg. Just as driving a spring-mass system at frequencies different from the resonance frequency can be energy-intensive, studies have shown that when animals bounce, they tend to maintain a certain stride frequency \citep{heglund1988speed}, below which locomotion can become more energy-intensive \citep{BLICKHAN19891217}. In line with these observations, numerous research studies have drawn an analogy between quadrupedal bouncing and a spring-loaded inverted pendulum (SLIP). Furthermore, in robotics, state-of-the-art approaches for designing bouncing gaits in quadruped robots often model the robot as a reduced-order SLIP model to mitigate the computational complexity that the entire high-dimensional robot model could entail \citep{ma2019first}. In this paper, we also employ a reduced-order SLIP-based model of quadruped dynamics. Building on this abstraction, we develop a bio-inspired controller that leverages energy conservation principles to generate stable and efficient bouncing gaits. Our approach integrates energy-aware trajectory synthesis with a feedback cancellation strategy, resulting in a lightweight and computationally efficient control scheme. This enables robust performance in the face of the model's inherent nonlinearity, hybrid dynamics, and potential state estimation errors.

\section{Background}
The seminal work of \citet{raibert1986legged} in the 1980s demonstrated that relatively simple control strategies could stabilize high-speed locomotion in legged robots, including quadrupeds. Raibert's model captured key characteristics later formalized in the spring-loaded inverted pendulum (SLIP) framework \citep{geyer2018gait}, which has since become a cornerstone for studying and designing dynamic legged systems due to its simplicity and computational efficiency.

Over the years, significant research has focused on aligning robotic mechanical designs with the SLIP model's principles \citep{gregorio1997design, 1705587, 677072}. While biological quadrupeds exhibit anatomical structures that differ from the idealized SLIP model, their center of mass (COM) dynamics often approximate SLIP-like behaviors during running \citep{geyer2018gait}. This insight has motivated the development of more complex robotic platforms that integrate SLIP dynamics at the control level, rather than strictly at the mechanical design level \citep{saranli2001rhex, cham2002fast}. 

In this work, we assume that the SLIP dynamics is embedded in the robot, whether through mechanical design or through any of the low-level control methods that have been extensively developed in previous works \citep{geyer2018gait}. Our focus is, thus, on designing a control strategy that achieves stable and robust locomotion within this reduced-order framework.

\input{tikz/SLIP}

Previous research on quadruped gait stabilization can generally be categorized into two main approaches: \textit{static} \citep{hugel1999towards, hardarson2002stability, ma2005omnidirectional, bai1999new} and \textit{dynamic} \citep{poulakakis2005modeling, kolter2008control}. While static gaits—such as walking and crawling—have been extensively studied, dynamic gaits like trotting and running have received comparatively less focus, primarily due to the added complexity of maintaining dynamic balance and managing transitions between flight and stance phases \citep{marhefka2003intelligent, kolter2008control}. 

This work focuses on the stabilization of dynamic gaits in quadrupeds. Earlier approaches to this problem typically employed linear control methods, whereas more recent efforts have increasingly adopted optimization-based techniques \citep{marhefka2003intelligent, zhang2018mode, saud/stability-tractability}. Although these methods offer improved performance in some contexts, they often entail high computational costs, which can limit their applicability in real-time scenarios on systems with limited resources.

In this paper, we introduce a lightweight, feedback cancellation-based tracking controller for quadruped robots exhibiting SLIP-like dynamics. The controller tracks an energy-aware, parabolic reference trajectory generated during the stance phase, enabling robust and efficient locomotion even in the face of measurement noise and system uncertainties. We validate the proposed method through simulation studies that demonstrate its ability to achieve stable, repeatable locomotion with low computational overhead.

\section{Contributions}
The key contributions of this paper are as follows: 
\begin{enumerate} 
    \item We propose a bio-inspired and computationally efficient control strategy for quadruped locomotion, based on energy conservation principles and reduced-order SLIP dynamics. 
    \item We employ a feedback cancellation-based approach to track a parabolic reference trajectory derived from energy conservation, and we use it to demonstrate stable and robust bouncing gaits. 
    \item We validate our approach through MATLAB simulations based on the physical parameters of the Ghost Robotics Minitaur\textsuperscript{\texttrademark} \citep{7403902}, showing reliable performance under nominal conditions and robustness to state measurement errors of up to 10\%.
\end{enumerate}

\section{SLIP Model of Quadruped Bouncing} \label{sec:slip}

Given the complexity of modeling and controlling full-order quadrupedal dynamics--which involve nonlinearities, hybrid transitions, and high dimensionality, we consider a reduced-order SLIP model. The SLIP model captures the fundamental mechanics of running as a repetitive cycle of ballistic motion and elastic rebound, while substantially reducing both the analytical and computational burden relative to full-order quadruped models. In practice, such SLIP-like dynamics can be incorporated into a quadruped's center-of-mass (COM) behavior either through mechanical design or via existing low-level control strategies.

\input{tikz/energy_conv}

\begin{figure}[h]
    \centering

\tikzset{every picture/.style={line width=0.75pt}} %set default line width to 0.75pt        

\begin{tikzpicture}[x=0.75pt,y=0.75pt,yscale=-1,xscale=1]

%Curve Lines [id:da8563852318071076] 
\draw [color={rgb, 255:red, 0; green, 0; blue, 0 }  ,draw opacity=0.5 ][line width=1.5]  [dash pattern={on 1.69pt off 2.76pt}]  (179.17,72.58) .. controls (272.67,143.58) and (384.67,124.58) .. (451.17,72.58) ;
%Shape: Circle [id:dp8495246644830867] 
\draw  [draw opacity=0][fill={rgb, 255:red, 255; green, 255; blue, 255 }  ,fill opacity=1 ] (369.82,104.18) .. controls (369.82,93.92) and (378.14,85.6) .. (388.4,85.6) .. controls (398.67,85.6) and (406.99,93.92) .. (406.99,104.18) .. controls (406.99,114.45) and (398.67,122.77) .. (388.4,122.77) .. controls (378.14,122.77) and (369.82,114.45) .. (369.82,104.18) -- cycle ;
%Shape: Circle [id:dp9953190999548139] 
\draw  [fill={rgb, 255:red, 255; green, 255; blue, 255 }  ,fill opacity=1 ] (296.46,117.97) .. controls (296.46,107.71) and (304.78,99.39) .. (315.04,99.39) .. controls (325.3,99.39) and (333.62,107.71) .. (333.62,117.97) .. controls (333.62,128.24) and (325.3,136.56) .. (315.04,136.56) .. controls (304.78,136.56) and (296.46,128.24) .. (296.46,117.97) -- cycle ;
%Shape: Circle [id:dp5418272411335259] 
\draw  [fill={rgb, 255:red, 255; green, 255; blue, 255 }  ,fill opacity=1 ] (224.06,102.18) .. controls (224.06,91.92) and (232.38,83.6) .. (242.64,83.6) .. controls (252.9,83.6) and (261.22,91.92) .. (261.22,102.18) .. controls (261.22,112.45) and (252.9,120.77) .. (242.64,120.77) .. controls (232.38,120.77) and (224.06,112.45) .. (224.06,102.18) -- cycle ;
%Straight Lines [id:da7232366780141731] 
\draw    (201,200.5) -- (440,200.5) ;
%Shape: Circle [id:dp35669202521953947] 
\draw  [fill={rgb, 255:red, 74; green, 144; blue, 226 }  ,fill opacity=0.57 ][line width=1.5]  (296.44,117.95) .. controls (296.45,107.69) and (304.78,99.38) .. (315.04,99.39) .. controls (325.3,99.4) and (333.61,107.73) .. (333.6,117.99) .. controls (333.59,128.26) and (325.26,136.57) .. (315,136.56) .. controls (304.74,136.55) and (296.43,128.22) .. (296.44,117.95) -- cycle ;
%Shape: Resistor [id:dp623354170447577] 
\draw  [line width=1.5]  (314.68,143.58) -- (314.68,151.65) -- (322.01,153.45) -- (307.34,157.02) -- (322,160.62) -- (307.33,164.19) -- (321.99,167.79) -- (307.32,171.36) -- (321.98,174.96) -- (307.31,178.53) -- (314.64,180.33) -- (314.64,188.4) ;
%Straight Lines [id:da9359976922845358] 
\draw [line width=1.5]    (314.64,188.4) -- (314.62,200.23) ;
%Shape: Circle [id:dp03065681237067508] 
\draw  [fill={rgb, 255:red, 74; green, 144; blue, 226 }  ,fill opacity=0.57 ][line width=1.5]  (227.59,113.09) .. controls (221.57,104.78) and (223.42,93.16) .. (231.73,87.14) .. controls (240.04,81.11) and (251.66,82.97) .. (257.68,91.27) .. controls (263.71,99.58) and (261.86,111.2) .. (253.55,117.23) .. controls (245.24,123.25) and (233.62,121.4) .. (227.59,113.09) -- cycle ;
%Shape: Resistor [id:dp09929190847908342] 
\draw  [line width=1.5]  (252.9,118.5) -- (262.32,131.5) -- (270.35,130.08) -- (262.67,144.47) -- (278.73,141.64) -- (271.04,156.02) -- (287.11,153.19) -- (279.42,167.58) -- (295.48,164.74) -- (287.8,179.13) -- (295.83,177.71) -- (305.25,190.71) ;
%Straight Lines [id:da24033368581623715] 
\draw [line width=1.5]    (305.25,190.71) -- (312.2,200.29) ;
%Shape: Circle [id:dp014089621047626899] 
\draw  [color={rgb, 255:red, 0; green, 0; blue, 0 }  ,draw opacity=0.2 ][fill={rgb, 255:red, 74; green, 144; blue, 226 }  ,fill opacity=0.2 ][line width=1.5]  (373.38,93.24) .. controls (379.42,84.95) and (391.04,83.12) .. (399.34,89.16) .. controls (407.64,95.2) and (409.47,106.82) .. (403.43,115.12) .. controls (397.38,123.42) and (385.76,125.25) .. (377.46,119.2) .. controls (369.17,113.16) and (367.34,101.54) .. (373.38,93.24) -- cycle ;
%Shape: Resistor [id:dp8537262178320705] 
\draw  [color={rgb, 255:red, 0; green, 0; blue, 0 }  ,draw opacity=0.1 ][line width=1.5]  (376.05,118.98) -- (366.6,131.96) -- (370.43,139.16) -- (354.38,136.3) -- (362.03,150.7) -- (345.98,147.83) -- (353.63,162.23) -- (337.57,159.37) -- (345.23,173.77) -- (329.17,170.91) -- (333,178.11) -- (323.55,191.09) ;
%Straight Lines [id:da08226937619348929] 
\draw [color={rgb, 255:red, 0; green, 0; blue, 0 }  ,draw opacity=0.2 ][line width=1.5]    (323.55,191.09) -- (316.59,200.65) ;
%Straight Lines [id:da6695849735630727] 
\draw [line width=1.5]    (314.68,143.58) -- (314.68,136.58) ;
%Shape: Circle [id:dp7532777344265216] 
\draw  [fill={rgb, 255:red, 0; green, 0; blue, 0 }  ,fill opacity=1 ] (353.57,115.18) .. controls (353.57,113.62) and (354.84,112.35) .. (356.4,112.35) .. controls (357.97,112.35) and (359.24,113.62) .. (359.24,115.18) .. controls (359.24,116.75) and (357.97,118.02) .. (356.4,118.02) .. controls (354.84,118.02) and (353.57,116.75) .. (353.57,115.18) -- cycle ;
%Straight Lines [id:da7138309628260276] 
\draw    (316.59,200.65) -- (352.74,122.39) ;
\draw [shift={(354,119.67)}, rotate = 114.8] [fill={rgb, 255:red, 0; green, 0; blue, 0 }  ][line width=0.08]  [draw opacity=0] (8.93,-4.29) -- (0,0) -- (8.93,4.29) -- cycle    ;
%Curve Lines [id:da11907769618311703] 
\draw    (349,199.67) .. controls (351.9,183.26) and (346.41,178.5) .. (334.34,166.5) ;
\draw [shift={(333,165.17)}, rotate = 45] [color={rgb, 255:red, 0; green, 0; blue, 0 }  ][line width=0.75]    (10.93,-3.29) .. controls (6.95,-1.4) and (3.31,-0.3) .. (0,0) .. controls (3.31,0.3) and (6.95,1.4) .. (10.93,3.29)   ;
%Straight Lines [id:da28511651224252454] 
\draw    (218.33,180.5) -- (218.33,145.17) ;
\draw [shift={(218.33,143.17)}, rotate = 90] [color={rgb, 255:red, 0; green, 0; blue, 0 }  ][line width=0.75]    (10.93,-3.29) .. controls (6.95,-1.4) and (3.31,-0.3) .. (0,0) .. controls (3.31,0.3) and (6.95,1.4) .. (10.93,3.29)   ;
%Straight Lines [id:da03557754264520718] 
\draw    (212.67,174) -- (247.33,174) ;
\draw [shift={(249.33,174)}, rotate = 180] [color={rgb, 255:red, 0; green, 0; blue, 0 }  ][line width=0.75]    (10.93,-3.29) .. controls (6.95,-1.4) and (3.31,-0.3) .. (0,0) .. controls (3.31,0.3) and (6.95,1.4) .. (10.93,3.29)   ;
%Curve Lines [id:da12629252943923974] 
\draw    (340,74.33) .. controls (360,81.33) and (342,94.33) .. (353,107.33) ;

% Text Node
\draw (331.4,138.75) node [anchor=north west][inner sep=0.75pt]    {$r$};
% Text Node
\draw (351.6,173.87) node [anchor=north west][inner sep=0.75pt]    {$\theta $};
% Text Node
\draw (227,176.4) node [anchor=north west][inner sep=0.75pt]    {$x$};
% Text Node
\draw (205,158.07) node [anchor=north west][inner sep=0.75pt]    {$y$};
% Text Node
\draw (282,67) node [anchor=north west][inner sep=0.75pt]   [align=left] {COM ($\displaystyle m$)};

\end{tikzpicture}

\caption{Depiction of a SLIP model during stance phase. Here $r$ measures the distance from the foot to the center of mass, and $\theta$ measures the angular displacement w.r.t. the ground.}
\label{fig:stance}
\end{figure}
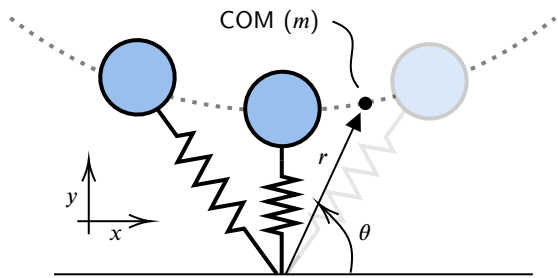

The SLIP model abstracts the quadruped as a point mass $m$ bouncing on a massless, compliant leg (Figure \ref{fig:slip}). It models the core energy exchanges that characterize running: kinetic energy is temporarily converted into elastic potential energy as the leg compresses during stance, then released to propel the body forward during takeoff. The model formalizes this behavior as a hybrid system alternating between a \textit{flight phase}, where the quadruped follows a ballistic trajectory, and a \textit{stance phase}, where the leg interacts with the ground as a spring-like mechanism, regulating both energy storage and release:
\[ \textbf{Flight:}\quad
    \begin{bmatrix} 
    m\ddot{x}\\
    m\ddot{y}
    \end{bmatrix}
    =
    \begin{bmatrix} 
    0\\
    -mg
    \end{bmatrix}
    ,
\]
\[ \textbf{Stance:}\quad
    \begin{bmatrix}
    m\ddot{r}\\
    \cfrac{d}{dt} \, m r^2 \dot{\theta}
    \end{bmatrix}
    =
    \begin{bmatrix}
    m r \dot{\theta}^2 - m g \sin \theta - b \dot{r} - \cfrac{d}{dr} \, U(r)\\
    m g r \cos \theta
    \end{bmatrix}.\\\vspace{10pt}
\] 
Here, $(x,y)$ is the position of the point mass $m$ in flight (see Figure \ref{fig:energy-conv}), and $(r, \theta)$ is its position in stance (see Figure \ref{fig:stance}). $U(r)$ is the general spring potential, which describes the radial force that the spring exerts on the point mass. For a linear spring with stiffness $k$ and rest length $r_0$,
\[
U(r) = \frac{1}{2} k \left( r_0 - r \right)^2.
\]

In our model, we assume control authority is limited to leg-length modulation during stance and leg-orientation adjustment during flight. This is a practical and widely adopted assumption, as most quadruped robots are equipped with actuated springs for leg-length control and servo motors for leg orientation during the aerial phase. This control structure mirrors biological quadrupeds, which regulate their body orientation in flight by shifting mass and adjust leg length through folding and extension. While animals also modulate leg stiffness during locomotion, we omit this behavior in our model, as most robotic platforms lack the mechanical complexity required for active stiffness control \citep{gregorio1997design}.

% We chose vertical height of the center-of-mass (COM) as an indicator of periodic hopping motion. Therefore, it would be of interest to see the phase portrait of the height as a state. This is portrayed in Figure \ref{fig:phase}. We provide a demonstrative video of the single-leg stance transitioning to flight at \url{https://youtu.be/_MZBhXTWy_k}.

\subsection{Stability Analysis}
The stability of periodic dynamical systems, such as the SLIP model, can be formally analyzed using Poincaré return maps. These discrete maps reduce the study of an $n$-dimensional continuous system (with $n = 4$ in our case) to the analysis of a discrete system on an $(n-1)$-dimensional subspace of the state space. While a rigorous Poincaré-based stability analysis is beyond the scope of this paper, we validate the stability properties of our system through MATLAB simulations.

\input{tikz/controller_overview}

\section{Energy-Conservation-Based Controller} \label{sec:controller}
Modeling the robot's center-of-mass (COM) as a reduced-order SLIP system enables us to focus on high-level control to regulate gait. An overview of our control architecture for this task is presented in Figure \ref{fig:controller-overview}. The inputs to the controller are the desired forward velocity and terrain information. The controller maintains the desired velocity by adjusting the leg's angle of attack during the flight phase, ensuring that appropriate propulsive forces are generated upon touchdown. Terrain information is used to compute the minimum apex height required for safe locomotion, which is then mapped, via energy conservation principles, into the target leg compression for the subsequent stance phase.

\subsection{Controller Architecture}
A detailed view of the controller is shown in Figure \ref{fig:controller-full}. The architecture consists of two core modules. During stance, a local controller regulates leg extension (based on the reduced-order SLIP model described in Section \ref{sec:slip}) to track a parabolic trajectory in flight that is derived using our energy-conserving framework, described below. An observer is added in loop with the system that attempts to estimate the states of the system based on its outputs. This emulates two important real-world behaviours: (1) it replicates the system's response in the presence of external noise, and (2) it emulates the system's response when confronted with incorrect state measurements from the sensors. 

Subsequently, once in flight, a servo controller takes charge, regulating the leg angle to a fixed angle of attack in preparation for the upcoming stance phase. Simultaneously, during flight, the system measures the current apex height and its difference from the desired height to calculate the energy required from the springs for the ensuing stance phase.

\input{tikz/controller}

\subsection{Constant Energy Formulation}
In our control design, we consider energy conservation over an entire gait cycle. This implies that, based on a reference energy value, our controller must introduce additional non-conservative work to maintain that energy level. However, given our formulation of the SLIP model, which only allows for energy injection into the system through spring actuation during stance phase, we need to design the controller so that it controls the behavior in stance that is reflected in flight phase. We borrow the idea from the selection of candidate Lyapunov functions in certain cases where the Hamiltonian (or total energy) of the system is a sufficient candidate function for local stability. If $\mathcal{H}_{i}$ is the Hamiltonian of the system during the $i^{th}$-phase of locomotion, then
\begin{align*}
    \mathcal{H}_{\text{fl}} &= \mathcal{H}_{\text{st}}\\
    W_{\text{fl}} + V_{\text{fl}} &= W_{\text{st}} + V_{\text{st}}\\
    V_{c_{\text{fl}}} + T_{\text{fl}} + W_{{nc}_{\text{fl}}} &= V_{c_{\text{st}}} + T_{\text{st}} + W_{{nc}_{\text{st}}}
\end{align*}
where $V_i$ is the potential energy, $T_i$ is the kinetic energy and $W_nc_i$ is the non-conservative work done (frictional loss, Rayleigh's dissipation, input energy, contact) during phase $i \in \{\text{fl}, \text{st}\}$  (where "fl" refers to the flight phase and "st" to the stance phase). Therefore, losses in energy can  be compensated for by injecting or removing energy as and when required.

Of particular interest along the trajectory are points where the potential energy is at its maximum: the apex height and the lowest stance point. At these instances, the system possesses maximum gravitational potential energy and maximum elastic potential energy, respectively. Additionally, the vertical velocity components vanish, and therefore, disregarding any non-conservative work done, there are only horizontal velocity components. Under this condition, the horizontal velocity components are the same. Should the objective be to reach a specific apex height during the flight phase for terrain navigation, the reference energy can be represented in terms of the required height and the vertical velocity. This makes it feasible to devise a trajectory that accomplishes these goals.

\subsection{Reference Trajectory Design}
The constant energy formulation enables us to design a reference trajectory that can then be tracked using a feedback cancellation-based controller. We begin by prescribing a takeoff angle that is equal in magnitude but opposite in direction to the landing angle, which is held fixed for each step. At this landing angle, we determine the $x$ and $y$ coordinates at which liftoff occurs--specifically when the leg spring reaches its nominal length. From this, we compute the horizontal distance between the takeoff and landing points, and subsequently, using the desired forward velocity of the robot, $v_h$, we calculate the time in stance as:

\[
    t = \frac{x}{v_h}
\]

Next, we calculate the final point on the parabolic stance trajectory. This is the point where the COM reaches its lowest height, and it is determined using conservation of energy as:

\[
    \Delta y_{\text{min}} = \sqrt{\frac{2mg\Delta h_{\text{apex}}}{k}}
\]

This equation holds because we enforce a constant horizontal velocity, along with the fact that vertical velocity is zero at both the apex of flight and the lowest stance point.

With the expected landing point $(x_0, y_0)$, lowest point in stance $(x_1, y_1)$, and takeoff point $(x_2, y_2)$ determined, along with the stance duration $t$, we formulate a time-parameterized ballistic trajectory for the COM that can be tracked with a linear controller. This trajectory is a quadratic of the form:
\[y(t)=ax(t)^2+bx(t)+c,\] satisfying the constraints
\begin{align*}
    y_0 &= ax_0^2 + bx_0 + c,\\
    y_1 &= ax_1^2 + bx_1 + c,\\
    y_2 &= ax_2^2 + bx_2 + c.
\end{align*}
The above system can be expressed in matrix form:
\[
    \underbrace{\begin{bmatrix}
    y_0\\[0.3em] y_1 \\[0.3em] y_2
    \end{bmatrix}}_{\mathbf{y}}
    = \underbrace{\begin{bmatrix}
    x_0^2 & x_0 & 1\\[0.3em]
    x_1^2 & x_1 & 1\\[0.3em]
    x_2^2 & x_2 & 1
    \end{bmatrix}}_{A}
    \underbrace{\begin{bmatrix}
    a \\[0.3em] b \\[0.3em] c
    \end{bmatrix}}_{\mathbf u}.
\]
Since $\mathbf y$ and $A$ are known, the matrix equation $\mathbf y = A \mathbf u$ can be solved for the vector $\mathbf u$, which contains the coefficients $a$, $b$, and $c$ of the quadratic $y(t)$. Once this quadratic trajectory is known, it is tracked using the feedback cancellation control law formulated in the next section.

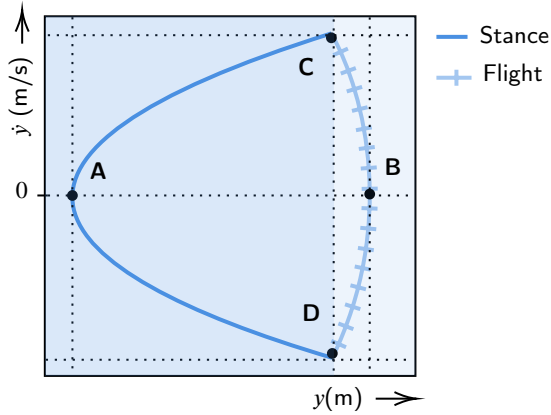
\begin{figure}[h!]
    \centering

\tikzset{every picture/.style={line width=0.75pt}} %set default line width to 0.75pt        

\begin{tikzpicture}[x=0.75pt,y=0.75pt,yscale=-1,xscale=1]

%Shape: Rectangle [id:dp07534062901458038] 
\draw   (200,47.59) -- (385,47.59) -- (385,227.1) -- (200,227.1) -- cycle ;
%Curve Lines [id:da8073435819464008] 
\draw [color={rgb, 255:red, 74; green, 144; blue, 226 }  ,draw opacity=1 ][line width=1.5]    (343.5,218.08) .. controls (193.7,173.74) and (148.9,116.81) .. (343.03,56.14) ;
%Curve Lines [id:da7375446472823071] 
\draw [color={rgb, 255:red, 74; green, 144; blue, 226 }  ,draw opacity=0.5 ][line width=1.5]    (343.03,56.14) .. controls (360.46,87.26) and (375.39,152.43) .. (343.5,218.08)(351.29,63.71) -- (344,66.99)(355.24,73.42) -- (347.73,76.18)(358.37,82.82) -- (350.7,85.12)(361.09,92.98) -- (353.31,94.82)(363.18,103.04) -- (355.31,104.45)(364.68,112.82) -- (356.74,113.82)(365.69,123.02) -- (357.71,123.6)(366.16,133.61) -- (358.16,133.75)(366.07,143.62) -- (358.08,143.34)(365.44,153.88) -- (357.47,153.17)(364.21,164.36) -- (356.29,163.21)(362.53,174.05) -- (354.69,172.48)(360.3,183.86) -- (352.55,181.88)(357.48,193.78) -- (349.85,191.37)(354.02,203.76) -- (346.53,200.94)(350.34,212.79) -- (343.01,209.59) ;
%Straight Lines [id:da9306583714873671] 
\draw [color={rgb, 255:red, 74; green, 144; blue, 226 }  ,draw opacity=1 ][line width=1.5]    (395.47,57.86) -- (411,57.86) ;
%Straight Lines [id:da5373607716770514] 
\draw [color={rgb, 255:red, 74; green, 144; blue, 226 }  ,draw opacity=0.5 ][line width=1.5]    (395.47,78.26) -- (413,78.26) (405.47,74.26) -- (405.47,82.26) ;
%Straight Lines [id:da5544731084738467] 
\draw    (196.28,137.03) -- (201.49,137.03) ;
%Straight Lines [id:da3964317688413598] 
\draw    (365.16,238.46) -- (383.46,238.46) ;
\draw [shift={(385.46,238.46)}, rotate = 180] [color={rgb, 255:red, 0; green, 0; blue, 0 }  ][line width=0.75]    (10.93,-3.29) .. controls (6.95,-1.4) and (3.31,-0.3) .. (0,0) .. controls (3.31,0.3) and (6.95,1.4) .. (10.93,3.29)   ;
%Straight Lines [id:da7857412659574765] 
\draw    (188.69,61.24) -- (188.69,46.29) ;
\draw [shift={(188.69,44.29)}, rotate = 90] [color={rgb, 255:red, 0; green, 0; blue, 0 }  ][line width=0.75]    (10.93,-3.29) .. controls (6.95,-1.4) and (3.31,-0.3) .. (0,0) .. controls (3.31,0.3) and (6.95,1.4) .. (10.93,3.29)   ;
%Shape: Ellipse [id:dp29513061345774716] 
\draw  [fill={rgb, 255:red, 0; green, 0; blue, 0 }  ,fill opacity=1 ] (211.59,137.03) .. controls (211.59,135.83) and (212.57,134.85) .. (213.78,134.85) .. controls (214.98,134.85) and (215.96,135.83) .. (215.96,137.03) .. controls (215.96,138.24) and (214.98,139.22) .. (213.78,139.22) .. controls (212.57,139.22) and (211.59,138.24) .. (211.59,137.03) -- cycle ;
%Shape: Ellipse [id:dp04480538422003644] 
\draw  [fill={rgb, 255:red, 0; green, 0; blue, 0 }  ,fill opacity=1 ] (359.99,136.33) .. controls (359.99,135.13) and (360.97,134.15) .. (362.18,134.15) .. controls (363.38,134.15) and (364.36,135.13) .. (364.36,136.33) .. controls (364.36,137.54) and (363.38,138.52) .. (362.18,138.52) .. controls (360.97,138.52) and (359.99,137.54) .. (359.99,136.33) -- cycle ;
%Straight Lines [id:da8420189767226793] 
\draw  [dash pattern={on 0.84pt off 2.51pt}]  (201.49,137.03) -- (384,137.03) ;
%Straight Lines [id:da5231617129980325] 
\draw  [dash pattern={on 0.84pt off 2.51pt}]  (213.78,48.03) -- (213.78,226.03) ;
%Straight Lines [id:da21520870818915017] 
\draw  [dash pattern={on 0.84pt off 2.51pt}]  (362.18,48.15) -- (362.18,226.15) ;
%Straight Lines [id:da27865220723360384] 
\draw  [dash pattern={on 0.84pt off 2.51pt}]  (201.49,57.03) -- (384,57.03) ;
%Straight Lines [id:da2550510479676946] 
\draw  [dash pattern={on 0.84pt off 2.51pt}]  (200.49,219.03) -- (383,219.03) ;
%Shape: Ellipse [id:dp19974159315228146] 
\draw  [fill={rgb, 255:red, 0; green, 0; blue, 0 }  ,fill opacity=1 ] (340.85,58.33) .. controls (340.85,57.12) and (341.83,56.14) .. (343.03,56.14) .. controls (344.24,56.14) and (345.22,57.12) .. (345.22,58.33) .. controls (345.22,59.54) and (344.24,60.52) .. (343.03,60.52) .. controls (341.83,60.52) and (340.85,59.54) .. (340.85,58.33) -- cycle ;
%Shape: Ellipse [id:dp09315792360918629] 
\draw  [fill={rgb, 255:red, 0; green, 0; blue, 0 }  ,fill opacity=1 ] (341.31,215.89) .. controls (341.31,214.68) and (342.29,213.7) .. (343.5,213.7) .. controls (344.71,213.7) and (345.69,214.68) .. (345.69,215.89) .. controls (345.69,217.1) and (344.71,218.08) .. (343.5,218.08) .. controls (342.29,218.08) and (341.31,217.1) .. (341.31,215.89) -- cycle ;
%Straight Lines [id:da36806306598932514] 
\draw  [dash pattern={on 0.84pt off 2.51pt}]  (344.18,49.15) -- (344.18,227.15) ;
%Shape: Rectangle [id:dp7323067647575262] 
\draw  [draw opacity=0][fill={rgb, 255:red, 74; green, 144; blue, 226 }  ,fill opacity=0.2 ] (200,47.59) -- (344.18,47.59) -- (344.18,227.15) -- (200,227.15) -- cycle ;
%Shape: Rectangle [id:dp7377650046374418] 
\draw  [draw opacity=0][fill={rgb, 255:red, 74; green, 144; blue, 226 }  ,fill opacity=0.1 ] (344.18,47.59) -- (385,47.59) -- (385,227.1) -- (344.18,227.1) -- cycle ;

% Text Node
\draw (415.65,50.87) node [anchor=north west][inner sep=0.75pt]   [align=left] {Stance};
% Text Node
\draw (417.08,69.57) node [anchor=north west][inner sep=0.75pt]   [align=left] {Flight};
% Text Node
\draw (184.09,128.67) node [anchor=north west][inner sep=0.75pt]    {$0$};
% Text Node
\draw (180.54,109.18) node [anchor=north west][inner sep=0.75pt]  [rotate=-270]  {$\dot{y} \ \text{(m/s)}$};
% Text Node
\draw (332.07,231.23) node [anchor=north west][inner sep=0.75pt]    {$y\text{(m)}$};
% Text Node
\draw (221,119) node [anchor=north west][inner sep=0.75pt]   [align=left] {\textbf{A}};
% Text Node
\draw (368,117) node [anchor=north west][inner sep=0.75pt]   [align=left] {\textbf{B}};
% Text Node
\draw (325,67) node [anchor=north west][inner sep=0.75pt]   [align=left] {\textbf{C}};
% Text Node
\draw (327,190) node [anchor=north west][inner sep=0.75pt]   [align=left] {\textbf{D}};

\end{tikzpicture}

    \caption{Verticle dynamics of a SLIP model stabilized using our energy-conservation-based controller. The plot corresponds to multiple hops of the robot, but since the gait is stable -- each hop traces the same trajectory as the previous, thus overwriting it. For the sake of brevity, we have edited the plot to remove the tickmarks, and placed markers (\textbf{A}, \textbf{B}, \textbf{C} and \textbf{D}) to identify the important points on the trajectories. In particular, the markers labeled \textbf{A} and \textbf{B} denote the apex points during stance and flight phases, respectively. The marker labeled \textbf{C} identifies the liftoff event, where the robot becomes airborne. The marker labeled \textbf{D} identifies the touchdown event, where the robot lands back on the ground. In the complete phase plot in $\mathbb R^4$, the points \textbf{C} and \textbf{D} are located on the guard surfaces representing the discrete dynamics that switches the continuous dynamical models between stance and flight.}
    \label{fig:phase}
\end{figure}

\subsection{Feedback Cancellation Based Tracking}
Let $\dot x = Ax + Bu$ define a linear dynamical system, and let us define $e_1 = x - \hat{x}$ as the deviation of the observed state $\hat{x}$ from the actual state $x$, and $e_2 = y_d - y$ as the difference between the output $y$ of the system and the reference signal $y_d$. Additionally, let us define $y=Cx$, where $C$ maps the system's state $x$ to the sensor space. Then, the rate at which the error $e_2$ evolves over time is given by
\begin{align} \label{ctrl_design_1}
    \dot{e}_2 = \dot{y}_d - C \dot{x}.
\end{align}

\noindent Here  $\dot{x}$ is the system dynamics. Since we lack full-state information, the controller must construct its own estimate of this dynamics: $\dot{x} = A \hat{x} + Bu$. In order to ensure that the output error dynamics is convergent, i.e., $\dot{e}_2 = K e_2,$ for $K \prec 0$, we impose
\begin{align} \label{ctrL_design_2}
    u = (CB)^{-1} \left( \dot{y}_d - CA \hat{x} + Ke_2 \right).    
\end{align}

\noindent Under this control law, the overall error dynamics can be formulated as the matrix equation:
\[
    \underbrace{\begin{bmatrix}
    \dot{e}_1 \\ \dot{e}_2
    \end{bmatrix}}_{\mathbf{\dot e}} = 
    \underbrace{\begin{bmatrix}
    A - K_e C & 0\\
    -CA & -CK
    \end{bmatrix}}_{E}
    \underbrace{\begin{bmatrix}
    e_1\\ e_2
    \end{bmatrix}}_{\mathbf e},
\]

\noindent where $K_e$ is the observer gain matrix. Note that $\mathbf{\dot e}$ is convergent if the real part of the eigenvalues of $E$ are negative. Furthermore, it can be seen that the eigenvalues of $E$ include the eigenvalues of both $A-K_e C$ and $-CK$. Thus, the original system is stable if both $A-K_e C$ and $-CK$ are stable. Pole placement methods to stabilize $A-K_e C$ are already well-studied and known. What remains is to design $K$ to arbitrarily place the poles of $-CK$. Since,
\[
    C = \begin{bmatrix}
    1 & 0  &0 &0\\
    0 &0 &1 &0
    \end{bmatrix}, \quad
    K = \begin{bmatrix}
        k_{11} & k_{12}\\
        k_{21} & k_{22}\\
        k_{31} & k_{32}\\
        k_{41} & k_{42}
    \end{bmatrix},
\]
\noindent we have
\[
    -CK = - \begin{bmatrix}
    k_{11} & k_{12}\\
    k_{31} & k_{32}
    \end{bmatrix}.
\]

\begin{figure*}
    \centering
    \includegraphics[width=0.7\textwidth]{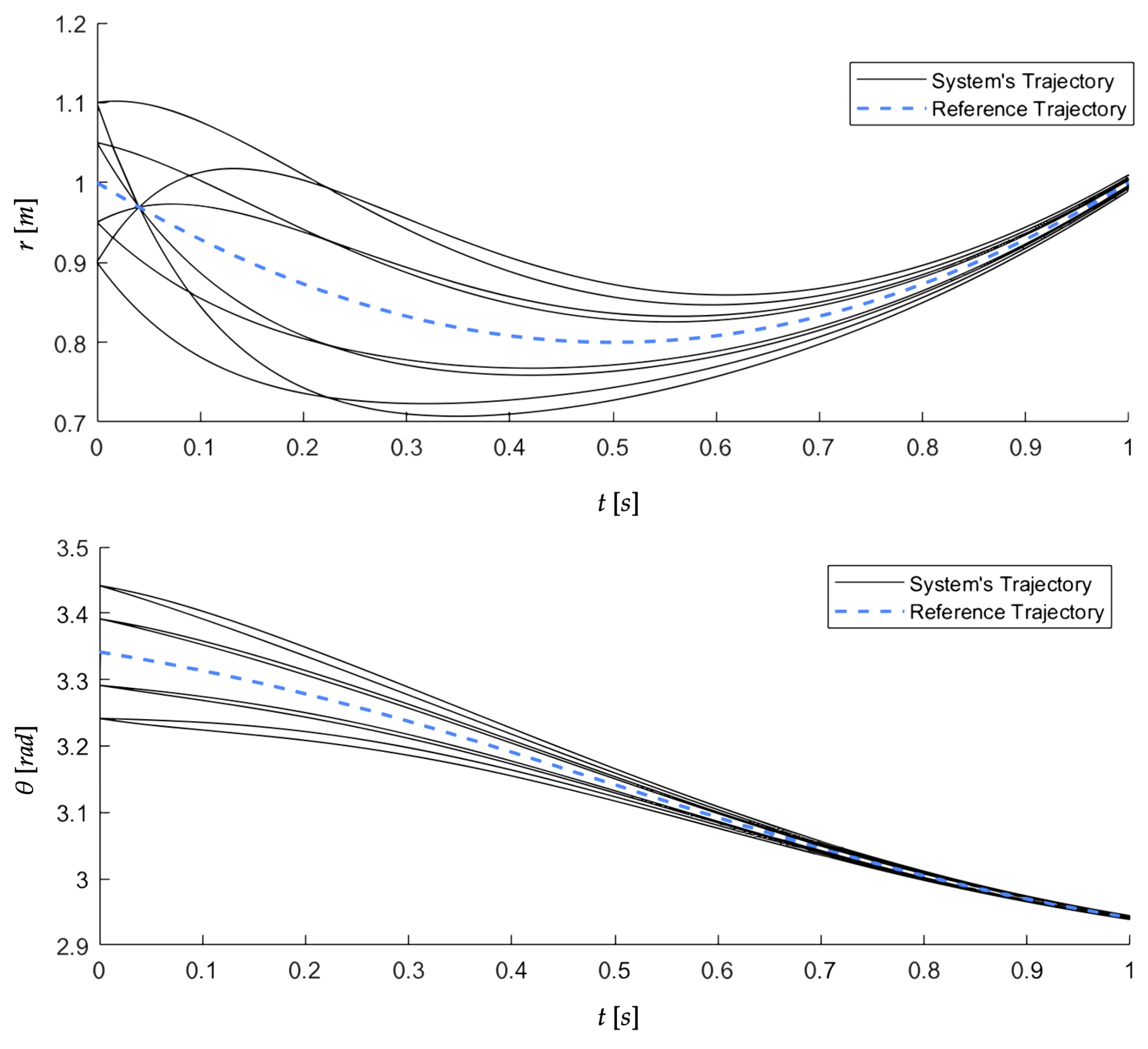}
    \caption{Response of the system with noisy touchdown measurements. It can be seen that even with a measurement error of up to 10\% in the touchdown state of a gait cycle, the controller is able to converge to the stable trajectory before the takeoff event. Thus, the controller is able to cancel out the error from potentially imprecise and noisy sensor measurements before the next gait cycle beings.}
    \label{fig:robust}
\end{figure*}

Notice that the entries $k_{21}, k_{22}, k_{41},$ and $k_{42}$ do not enter the expression for $-CK$, and thus can be chosen arbitrarily. Here, we will set them to $0$. Now, say we want the eigenvalues of $-CK$ to be $\eta_1$ and $\eta_2$. That is, we want 
\begin{align*}
    \det -CK - \lambda I &= (\lambda - \eta_1) (\lambda - \eta_2)\\
    (k_{11}+\lambda)(k_{32}+\lambda) - k_{31}k_{12} &= \lambda^2 - (\eta_1 + \eta_2) \lambda + \eta_1 \eta_2
\end{align*}
This is an overdetermined system of equations. Let us arbitrarily take $k_{32} = k_{31} = 1$ and solve the algebraic system for the rest:
\begin{align*}
    k_{11} &= -(\eta_1 + \eta_2 + 1),\\
    k_{12} &= -(\eta_1 + \eta_2 + \eta_1 \eta_2 + 1).
\end{align*}
Therefore, we have
\[
K = 
-\begin{bmatrix}
    \eta_1 + \eta_2 + 1 & \eta_1 + \eta_2 + \eta_1 \eta_2 + 1\\[0.3em]
    0 & 0\\[0.2em]
    -1 & -1\\[0.2em]
    0 & 0
\end{bmatrix}.
\]
    
Setting the controller gains $K$ as given above and the observer gains $K_e$ using standard pole placement, we can place the eigenvalues of $E$ arbitrarily and control the error dynamics $\mathbf e$. By placing the eigenvalues in the left-half of the complex plane, we can get the gains $K$ and $K_e$ that ensure $E \prec 0$, therefore making the error dynamics $\mathbf{\dot{e}} = E \mathbf e$ convergent.

\section{Results}
We demonstrate our energy-conservation-based controller on a SLIP model initialized using the mechanical parameters of a Ghost Robotics Minitaur\textsuperscript{\texttrademark} quadruped. In Figure \ref{fig:phase}, we present the $y$ vs. $\dot y$ slice of the system's phase space, which characterizes the vertical dynamics of the system. It can be seen that the vertical dynamics is stable: the system traces the exact same trajectory \textbf{A} $\to$ \textbf{B} $\to$ \textbf{C} $\to$ \textbf{D} in each gait cycle. 

In Figure \ref{fig:robust}, we present our controller's robustness to noisy state measurements. As seen, the controller is robust to up to 10\% noise or measurement error in estimates of the touchdown sensor. Therefore, our proposed control algorithm maintains stable bouncing dynamics despite noisy and imperfect sensor measurements. We provide a video depicting a robot bouncing with a stable gait under noisy touchdown measurements at \url{https://youtu.be/3NYftme75Qw}.

\section{Conclusion}
In this work, we developed a tracking controller for quadruped robots that generates stable and efficient gaits. Our approach leverages a bio-inspired energy conservation principle combined with a computationally efficient spring-loaded inverted pendulum (SLIP) model to capture the essential dynamics of quadrupedal locomotion. Through simulations using a model parameterized by the mechanical properties of the Ghost Robotics Minitaur\textsuperscript{\texttrademark}, we demonstrate that the controller consistently produces stable bouncing gaits and maintains robustness to state estimation errors of up to 10\%, highlighting its practical resilience to sensor noise and measurement uncertainties. Future work will focus on augmenting the controller with mechanisms to compensate for modeling inaccuracies and extending the energy-conserving parabolic formulation to 3D locomotion. Ultimately, we aim to validate the approach on a physical quadruped platform.

\section*{Data Access Statement}
No data were used in the conduct of this study.

\section*{Ethics Statement}
This study did not involve human or animal subjects, nor did it include any procedures or data that raise ethical concerns.

\section*{Acknowledgements}
No external funding was received for this work. The authors declare no conflicts of interest. Muhammad Saud Ul Hassan, Derek Vasquez, and Hamza Asif contributed equally to the research and writing of this manuscript.

\printcredits
\bibliographystyle{cas-model2-names}
\bibliography{cas-refs}

\end{document}